\documentclass[letterpaper]{article} 
\usepackage{aaai23}  
\usepackage{times}  
\usepackage{helvet}  
\usepackage{courier}  
\usepackage[hyphens]{url}  
\usepackage{graphicx} 
\def\UrlFont{\rm}  
\usepackage{natbib}  
\usepackage{caption} 
\usepackage{algorithm}
\usepackage{algorithmic}

\usepackage{newfloat}
\usepackage{listings}
\DeclareCaptionStyle{ruled}{labelfont=normalfont,labelsep=colon,strut=off} 
\floatstyle{ruled}
\newfloat{listing}{tb}{lst}{}
\floatname{listing}{Listing}
\usepackage{smile}

\newcommand{\ie}{\textit{i}.\textit{e}.}

\newcommand{\tabincell}[2]{\begin{tabular}{@{}#1@{}}#2\end{tabular}}  

\usepackage{color}
\definecolor{teal}{RGB}{99, 255, 172}

\title{Towards In-distribution Compatibility in Out-of-distribution Detection}
\author{
    Boxi Wu\textsuperscript{\rm 1}\equalcontrib, Jie Jiang\textsuperscript{\rm 2}\equalcontrib,
    Haidong Ren\textsuperscript{\rm 4},
    Zifan Du\textsuperscript{\rm 3},
    Wenxiao Wang\textsuperscript{\rm 3}, \\
    Zhifeng Li\textsuperscript{\rm 2},
    Deng Cai\textsuperscript{\rm 1},
    Xiaofei He\textsuperscript{\rm 1},
    Binbin Lin\textsuperscript{\rm 3},
    Wei Liu\textsuperscript{\rm 2}
}
\affiliations{
    \textsuperscript{\rm 1}State Key Lab of CAD\&CG, Zhejiang University.
    \textsuperscript{\rm 2}Tencent Data Platform. \\ 
    \textsuperscript{\rm 3} School of Software Technology, Zhejiang University.
    \textsuperscript{\rm 4}Ningbo Zhoushan Port Group Co.,Ltd., Ningbo, China. 

\nocopyright
}

\usepackage{bibentry}

\begin{document}

\maketitle

\begin{abstract}

Deep neural network, despite its remarkable capability of discriminating targeted in-distribution samples, shows poor performance on detecting anomalous out-of-distribution data. To address this defect, state-of-the-art solutions choose to train deep networks on an auxiliary dataset of outliers. Various training criteria for these auxiliary outliers are proposed based on heuristic intuitions. 
However, we find that these intuitively designed outlier training criteria can hurt in-distribution learning and eventually lead to inferior performance. To this end, we identify three causes of the in-distribution incompatibility: \textit{contradictory gradient}, \textit{false likelihood}, and \textit{distribution shift}.
Based on our new understandings, we propose a new out-of-distribution detection method by adapting both the top-design of deep models and the loss function. Our method achieves in-distribution compatibility by pursuing less interference with the probabilistic characteristic of in-distribution features. On several benchmarks, our method not only achieves the state-of-the-art out-of-distribution detection performance but also improves the in-distribution accuracy.

\end{abstract}

\section{Introduction}

Deep neural networks have achieved extraordinary performance across a wide range of artificial intelligence and pattern recognition tasks. Many of these tasks are formulated in a constrained scenario. That is, all the considered training and testing samples are assumed to belong to a few predefined limited categories. This is the case for many standard computer vision tasks such as classification~\cite{cifar}, detection, and segmentation. Naturally, people start to wonder how deep networks will react to out-of-distribution (OOD) data, data that do not belong to any of the predefined categories. 

\citet{ood-baseline} first studied this question and found that deep networks tend to assign high confidence scores to OOD samples. This problem hugely hampers safely deploying deep models in the open world. The behavior of artificial intelligence applications such as autonomous driving and medical image processing~\cite{RenLFSPDDL19} can be unpredictable when facing OOD data. Various approaches have been proposed to solve this problem. One of the most effective approaches is Outlier Exposure (OE)~\cite{oe}, which chooses to train deep networks with an auxiliary dataset of outliers. \cite{oe} pointed out that the more realistic the auxiliary dataset is, the better performance for OOD detection. 

The original work of OE adopts the KL-divergence to the uniform distribution as the training criterion of outliers.
Subsequent works followed the framework of Outlier Exposure and proposed various different training criteria of outliers for better effectiveness~\cite{energy}. Most of these are designed based on intuition. However, when experimenting with these methods, we find that training with OE can lead to inferior in-distribution accuracy. To better understand this issue, we dive into the detailed design of these OOD training criteria and identify three major factors that cause the in-distribution incompatibility:

\begin{figure}[t!]
\centering
\includegraphics[width=0.95\linewidth]{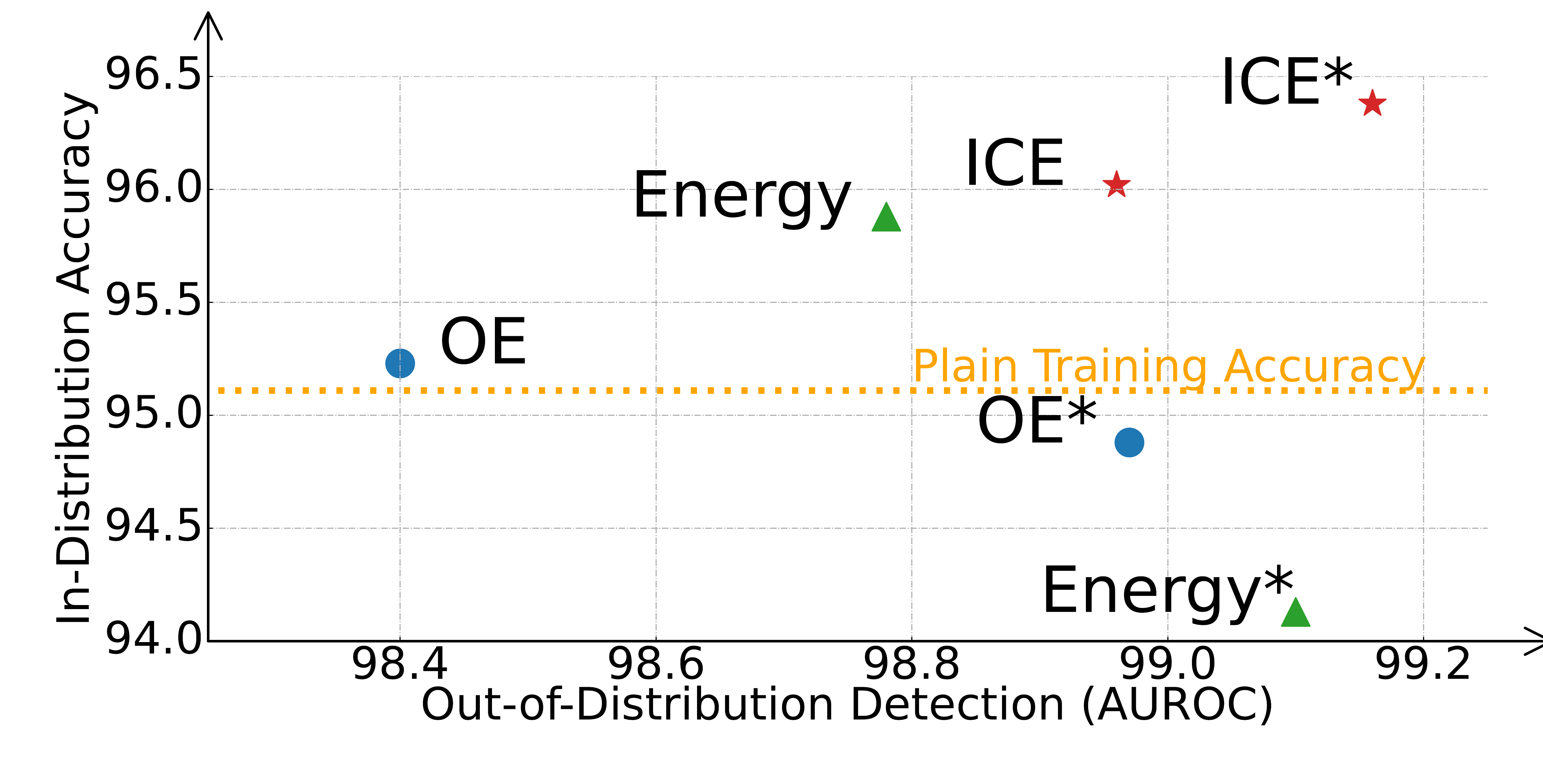}
\caption{The in-distribution accuracies and out-of-distribution detection AUROC scores of different models. Models annotated with * indicate the model is trained from scratch; otherwise, the model is fine-tuned. Our ICE achieves superior performances on both metrics.}
\label{fig:tradeoff}
\end{figure}

\begin{enumerate}
    \item \textbf{Contradictory Gradient.} The training objective of existing OE methods occasionally generates gradients in the opposite direction of those generated by the in-distribution objective, which further hampers the in-distribution discriminant learning.
    
    \item \textbf{False Likelihood.} Existing OE algorithms adopt the logit value generated by neural networks as a surrogate signal for likelihood estimation. Yet, under Gaussian discriminant analysis~\cite{mahala}, the logit value is a defective estimation of likelihood, which can result in assigning high in-distribution scores on outliers. 
    
    \item \textbf{Distribution Shift.} OE objective essentially shifts out-of-distribution features to the center of features space. During this process, in- and out-of-distribution features may mix with each other, which prevents separating different categories.
    
\end{enumerate}


These three factors are intrinsically related and together hamper in-distribution learning. In Section~3, we will demonstrate these three factors for both linear models as well as deep networks. Existing OE methods partially addressed one or two factors, intentionally or unintentionally. Yet, an in-distribution-compatible OE algorithm that can universally solve all these problems is still absent. Therefore, in Section~4, we intend to design a new OE method, \textbf{I}n-distribution \textbf{C}ompatible outlier \textbf{E}xposure (ICE), which can achieve high performance on in-distribution classification and out-of-distribution detection simultaneously, as shown in Fig.~\ref{fig:tradeoff}. 
Our loss function is designed based on the principle of \textbf{not contradicting the in-distribution gradients}.
Meanwhile, we focus on the probabilistic characteristic of in-distribution samples. Commonly-used deep models generate high-dimension in-distribution features that approximately form a class-conditional Gaussian distribution. Then a linear layer will transform the features to the scalar logit value. \textbf{To avoid the false likelihood problem}, we replace the linear layer with a Gaussian mixture model so that we can estimate the actual class-conditional likelihood. \textbf{To prevent distribution shift}, our loss function is designed to push the outliers away from the closest in-distribution cluster so that the in-distribution cluster can be minimally affected. 

We test ICE on benchmark datasets in Section~5. In the most challenging case where CIFAR10~\cite{cifar} serves as the in-distribution set and CIFAR100 as the out-of-distribution set, ICE improves the FPR95 score to $22.36\%$ with an improvement of $3.79\%$ over previous \textit{sota} results. Meanwhile, ICE achieves high in-distribution accuracy from $95.11\%$ to $96.38\%$ over the plain in-distribution training. The visualization of features learned by ICE further verifies the soundness of our design.

\section{Related Works}

The pioneering work of \citet{ood-baseline} pointed out the importance of detecting outliers with deep networks. \citet{relu} analyzed that the ReLU activation can hamper detecting outliers. To alleviate this issue, \citet{ood-baseline} proposed to use the maximum soft-max score as an indicator of outliers. Similarly, \citet{odin} proposed a refined outlier score by adopting temperature scaling and adding small perturbations to input. \citet{godin} further improved the framework of \citet{odin} by decomposing confidence scoring. Other different forms of confidence scores were kept being proposed \cite{RenLFSPDDL19,maxlogit}. \citet{mahala} considered using a confidence score of Mahalanobis distance to detect abnormal samples. \citet{in-n-out} address detection performance by using auxiliary information. 

The works mentioned above mainly improve detection performance without modifying the trained models. The methodology of Outlier Exposure~\cite{oe} considers using training techniques with an auxiliary dataset of outliers and achieves superior results. Many OOD detection techniques are constrained to small-scale datasets. Recently, several works~\cite{react,mos,udg} investigated the detection task on large-scale datasets such as ImageNet. \citet{mood} accelerated the detection speed with multi-level features. \citet{vim} improve the performance by crafting virtual logit from heterologous confidence scores. \citet{logit-norm} proposed an alternative loss function to replace the commonly-used softmax cross-entropy loss, which can be combined with outlier exposure methods for better performance.
\section{In-Distribution Incompatibility}

The state-of-the-art Outlier Exposure accomplishes the task of OOD detection via training outliers with an extra loss function. Different approaches propose their own outlier loss function. In this section, we use two of the most representative outlier loss functions, the standard KL-Divergence to uniform distribution~\cite{oe} and the Energy score~\cite{energy}, to illustrate the three causes of in-distribution incompatibility.

\subsection{Preliminaries}
\label{sec:preli}
We choose the fundamental problem of multi-class classification as the subject to illustrate our methodology. We first introduce the notations and rudiments of the studied topic as below.

\textbf{Notations.} For the multi-classification task with $K$ classes, a neural work will first transform input $\xb$ with a non-linear mapping to feature $\zb=Z(\xb) \in \RR^d$. $Z$ represents deep network backbone. Then, a linear model $f$ is supposed to output a real-value score for each class:
\begin{align}
\label{eq:linear}
f(\xb) = \bW \zb + \bb
.
\end{align} 
$\bW=[\bw_1, \cdots, \bw_K]$ and $\bb=[b_1, \cdots, b_K]$ are trainable parameters. $f_i(\zb) = \bw_i^\top \zb + b_i$ is the predicted score for class $i\in [K]$, which is known as the logit value. $[K]:=\{1, \cdots, K\}$. $\bw_i \in \RR^d$ and $b_i \in \RR$ compute a scalar score for class $i$ from feature $\zb$. Finally, the stochastic gradient-descending algorithm will optimize the Softmax Cross-Entropy loss:
\begin{align}
\mathcal{L}_\text{sce}(\xb,\bm{1}_y)
=
-
\bm{1}_y^\top
\log
\big[\softmax[f(\zb)]\big]
\label{eq:ce}
.
\end{align}
$y$ is the ground-truth label for $\xb$, and $\bm{1}_y \in \mathbb{R}^K$ is its one-hot encoding. We define the softmax function $\softmax(f) : \RR^K \to \RR^K$ as $\softmax(f)_i = \exp(f_i) / \sum_{k=1}^{K}\exp(f_k)$. the logarithm is defined as element-wise.

\textbf{Outlier Exposure.} Denoting the in-distribution set as $\cD_{in}$, any possible input that does not belong to $\cD_{in}$ is considered to belong to the OOD set $\cD_{out}$. The intriguing part of detecting OOD data is that we cannot cover the entire $\cD_{out}$ set during the training stage. Nevertheless, \citet{oe} showed that a good choice of a subset of $\cD_{out}$ is crucial for learning models that can effectively detect unseen testing OOD data $\cD_{out}^{test}$. Particularly, denoting the auxiliary set of outliers as $\cD_{out}^{oe}$, the network is trained with:
\begin{align}
\label{eq:oe}
\underbrace{
\EE_{(\xb,y)\sim\cD_{in}}
\cL_\text{sce}(\xb,\bm{1}_y)
}_{\text{in-distribution risk}}
+
\lambda \cdot
\underbrace{
\mathbb{E}_{\tilde{\xb}\sim\mathcal{D}_{out}^{oe}}
\mathcal{L}_\text{sce}(\tilde{\xb}, \ub)
}_{\text{outlier exposure}}
.
\end{align}
$\ub \in \RR^K$ is the uniform distribution over $K$ classes. Weight parameter $\lambda$ balances the two training objectives. 
For better detection performance, \citet{energy} proposed to use the energy score as an alternative loss function for outliers:
\begin{align}
\cL_{\text{energy}}(\tilde{\xb})
\label{eq:energy}
=
\log \big[
\sum\mathop{}_{\mkern-5mu k \in [K]}\exp(\bw_k^\top \tilde{\zb} + b_k)
\big].
\end{align}
$\tilde{\zb}=Z(\tilde{\xb})$.
Minimizing the above metric results in low confidence in all the in-distribution categories. To balance in-distribution energy, the negative energy of in-distribution features $-\cL_{\text{energy}}(\xb) $ are simultaneously optimized:
\begin{align}
\label{eq:energy-full}
&\mathbb{E}_{(\xb,y)\sim\mathcal{D}_{in}}
\mathcal{L}_\text{sce}(\xb,\bm{1}_y) \
+
\\
&
\lambda \cdot 
\big[
\mathbb{E}_{(\xb)\sim\mathcal{D}_{in}}
-
\cL_{\text{energy}}(\xb) 
+
\mathbb{E}_{\tilde{\xb}\sim\mathcal{D}_{out}^{oe}}
\cL_{\text{energy}}(\tilde{\xb})
\big]
.
\notag
\end{align}
In the following parts, we will discuss the differences between these two carefully-designed criteria and their common deficiencies.


\subsection{Contradictory Gradient} 
The gradient generated by training criteria can reveal how it influences trainable variables. For instance, the softmax cross-entropy in Eqn.~\eqref{eq:ce} generates gradients as:
\begin{align}
\frac{\partial \cL_{\text{sce}}(\xb,\bm{1}_y)}{\partial f_i(\zb)}
=
\begin{cases}
\frac{\exp[f_i(\zb)]}{\sum_{k \in [K]}\exp[f_k(\zb)]} -1 <0,
\ \text{if} \ y = i  ;\\
\frac{\exp[f_i(\zb)]}{\sum_{k \in [K]}\exp[f_k(\zb)]} > 0,
\quad \ \ \ \  \text{if} \ y \neq i .
\end{cases}
\notag
\end{align}
With the gradient-descending optimization, $\cL_{\text{sce}}$ will increase the logit score $f_i(\xb)$ when $i$ being the ground-truth class, otherwise decrease $f_i(\xb)$. 
Such design aligns with the belief that $f_i(\xb)$ is a good hint of estimating the confidence that $\xb$ belongs to category $i$. 
However, the OE objective in Eqn.~\eqref{eq:oe} contradicts the principle that only the ground-truth logit $f_y(\xb)$ is enlarged during training.
Specifically, the gradient generated by $\cL_{\text{sce}}(\tilde{\xb},\ub)$ in Eqn.~\eqref{eq:oe} is:
\begin{align}
\label{eq:ce-grad}
\frac{\partial \cL_{\text{sce}}(\tilde{\xb},\ub)}{\partial f_i(\tilde{\zb})}
=
K \cdot
\Big[
\frac{\exp[f_i(\tilde{\zb})]}{\sum_{k \in [K]}\exp[f_k(\tilde{\zb})]} -
\frac{1}{K}
\Big]
.
\end{align}
When the softmax probability $\softmax(f)_i$ exceeds $1/K$, $\cL_{\text{sce}}(\tilde{\xb},\ub)$ will generate positive gradient and thus surpass the value of $f_i(\tilde{\zb})$. This is reasonable as it reduces the confidence of predicting class $i$ on outlier $\tilde{\xb}$.
The problem is, when $\softmax(f)_i < 1/K$, $\cL_{\text{sce}}(\tilde{\xb},\ub)$ will generate negative gradient and increases the confidence of predicting class $i$. Although such a case only happens when the posterior is lower than $1/K$, it still violates the principle mentioned above and increases the probability of mistaking $\tilde{\xb}$ with an in-distribution class.
In contrast, the energy objective $\cL_{\text{energy}}(\tilde{\xb})$ in \eqref{eq:energy} obeys the principle by punishing $f_i(\tilde{\zb})$ for all classes:
\begin{align}
\label{eq:energy-od-grad}
\frac{\partial \cL_{\text{energy}}(\tilde{\xb})}{\partial f_i(\tilde{\zb})}
=
\frac{\exp[f_i(\tilde{\zb})]}{\sum_{k \in [K]}\exp[f_k(\tilde{\zb})]} 
> 0
.
\end{align}
Despite that the $\cL_{\text{energy}}$ is compatible with in-distribution learning, the energy score falls into the issue of contradictory gradient due to the negative in-distribution energy $-
\cL_{\text{energy}}(\xb) $ in \eqref{eq:energy}:
\begin{align}
\label{eq:energy-id-grad}
\frac{\partial -\cL_{\text{energy}}(\xb)}{\partial f_i(\zb)}
=
-
\frac{\exp[f_i(\zb)]}{\sum_{k \in [K]}\exp[f_k(\zb)]} 
< 0
.
\end{align}
$-\cL_{\text{energy}}(\xb)$ increase $f_i(\zb)$ for all classes instead of only the ground-truth, which contradicts the gradient of $\partial \cL_{\text{sce}}(\xb,\bm{1}_y)/\partial f_i(\zb)$ when $i\neq y$. The contradictory gradient confuses the learning process by sending signals in opposite directions, eventually leading to inferior performance.

\begin{figure}[t!]
		\centering
\resizebox{0.98\linewidth}{!}{
		\subfigure[Logit Value $f_i(\zb)$.]{
		\label{fig:1a}
			\includegraphics[width=0.45\linewidth]{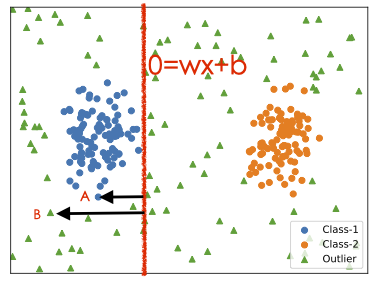}}

		\subfigure[Class Likelihood $p(i|\zb)$.]{
		\label{fig:1b}
			\includegraphics[width=0.45\linewidth]{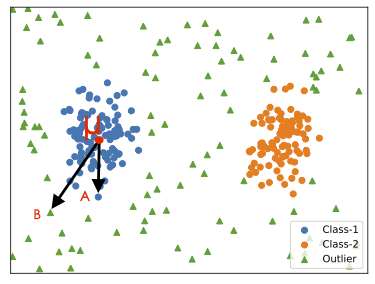}}
}
		\caption{In a 2-dimension binary classification problem, we compare the difference between logit value $f_i(\zb)$ and class likelihood $p(i|\zb)$. For in-distribution sample A and out-of-distribution sample B, $f_i(\zb)$ assigns higher confidence on B, while $p(i|\zb)$ assigns higher confidence on A.}
		\label{fig:1}
\end{figure}
\subsection{False Likelihood}

One effective way to detect outlier $\tilde{\xb}$ is by estimating the likelihood of each class. A potential outlier is supposed to have a low likelihood value for all classes. Many outlier loss functions are designed based on this belief. Yet, many of these loss functions use the logit value $f_i(\tilde{\zb})$ as a substitute for class likelihood $p(\xb|y=i)$. These two scores are not equivalent. Occasionally, an outlier with a low likelihood value $p(\xb|y=i)$ can be assigned with a high logit score.

\begin{figure*}[!t]
\centering
\includegraphics[width=\linewidth]{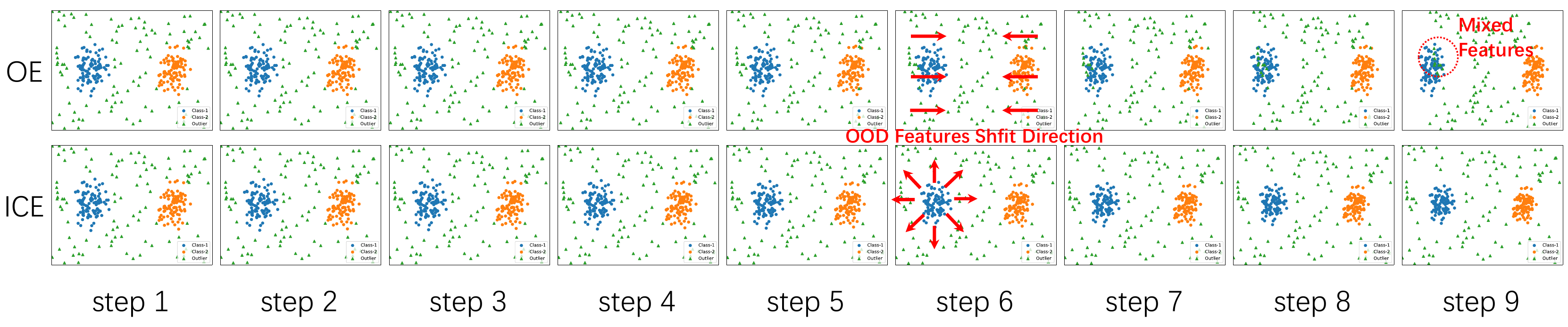}
\caption{The distribution shift effect is caused by different methods. We plot feature distribution for each step of SGD optimization. Standard Outlier Exposure shifts outliers to the center of space, during which the in- and out-of-distribution samples are mixed. On the other hand, ICE pushes outliers away from the class center and does not suffer from distribution shift.}
\label{fig:ds}
\end{figure*}

To understand this problem, we need to first inspect the intrinsic reasons why in-distribution learning uses the linear model in Eqn.~\eqref{eq:linear} to generate a confidence score $f_i$. 
Under the assumption that the learned features $\zb$ distribute as a class-conditional Gaussian: $p(\zb|y) \sim \cN (\bmu_y,\bSigma)$, $\bSigma$ is a tied covariance, and $\bmu_i$ is the mean of class $i$, the methodology of logistic regression deduces from the Bayes' Theorem that the posterior probability $p(i|\zb)$ is equal to:
\begin{align}
\notag
p(i|\zb)
=
\frac{{\hat{\bw}}_i^\top \zb + \hat{b_i}}{\sum_{k \in [K]}({\hat{\bw}}_k^\top \zb + \hat{b_k})}, \ \text{where}
\\
\hat{\bw}_i = \bSigma^{-1} \bmu_i, \
\hat{b}_i = - \frac{1}{2} \bmu_i^{\top}\bSigma^{-1} \bmu_i .
\end{align}
Thus, under the assumption that parameters $\bw_i$ and $b_i$ can fit the $\hat{\bw}_i$ and $\hat{b}_i$ from training, $f_i$ is a desired in-distribution discriminator. 
From geometric perspective, the linear model $f_i$ measures the distance to the hyper-plane of $\bm{0} = \bw_i^\top \zb + b_i$, while the likelihood $p(\zb|i)$ is proportional to the negative Mahalanobis distance to the class center $\bmu_i$. These two measurements are not equivalent. To demonstrate the difference, we visualize the case of a 2-dimensional binary classification problem. In Fig.~\ref{fig:1a}, we can observe that an in-distribution sample A has a lower value of $f_y$ compared with an out-of-distribution sample B, while the likelihood of A is higher than B. Such inconsistency between $f_i(\zb)$ and  $p(i|\zb)$ is overlooked by previous works, including OE and Energy.


\subsection{Distribution Shift} 

The high-dimension feature generated by the penultimate layer of the deep network, \ie $\zb=Z(\xb)$, is trainable. Thus, unlike the static $\xb$, the distribution of $\zb$ will shift with respect to the imposed loss functions. When only in-distribution features exist, the softmax cross-entropy loss $\cL_{\text{sce}}$ will push features away from the discriminant plane. However, when out-of-distribution features are considered, the shifted distribution between in-distribution features and out-of-distribution features may mix, which interferes with the discrimination of deep networks.

We still use the 2-dimension visualization for illustration.
In Fig.~2, we set the training features, either in-distribution or out-of-distribution, to be trainable to simulate features generated by deep networks and plot the distribution shift on each optimization step of the gradient-descending algorithm. Parameters $\bw_i$ and $b_i$ are initialized with $\hat{\bw}_i$ and $\hat{b}_i$.
The standard OE objective $\cL_{\text{sce}}(\tilde{\xb},\ub)$ will push the gradient to:
\begin{align}
\label{eq:logistic}
\partial \cL_{\text{sce}}(\tilde{\xb},\ub)/\partial \tilde{\zb}
=
&\big[2\cdot \softmax[f(\tilde{\zb})]_1 - 1 \big] \bw_1 
\\
+
&\big[2\cdot \softmax[f(\tilde{\zb})]_2 - 1 \big] \bw_2 
.
\notag
\end{align}
If $\softmax[f(\tilde{\zb})]_1 > 1/2$, the above gradient tend to decrease the value of $f(\tilde{\zb})_1$ and increases $f(\tilde{\zb})_2$, and vice versa if $\softmax[f(\tilde{\zb})]_2 > 1/2$.
As a result, out-of-distribution features are dragged to the space between the two in-distribution classes.
For OOD features that are initially far away from either of the two in-distribution classes, the OE objective may shift them to the position of in-distribution features during the optimization iterations. These misplaced features then confuse in-distribution learning, as the discriminator is required to make opposite predictions on similar features.
In the next section, we will show that our method avoids this issue by encouraging in-distribution features to gather in the center of its distribution and pushing outliers away from the center, as also shown in Fig.~\ref{fig:ds}.

\section{Methodology}

In this section, we introduce our \textbf{I}n-distribution \textbf{C}ompatible outlier \textbf{E}xposure (ICE) to address the three problems discussed before. To this end, we adapt both the top design of deep models and the loss function. Besides the refined in-distribution compatibility, our ICE also provides direct probabilistic confidence estimation and saves the usage of hyper-parameters.

\subsection{Modified Top-Design}
We first replace the traditional linear layer of deep networks with the class-conditional Gaussian model. Namely, instead of using  Eqn.~\eqref{eq:linear} to generate a confidence score, we opt for the following estimation based on Mahalanobis distance:
\begin{align}
h_i(\zb) = -
 (\zb-\mb_i )^{\top}
 (\Lb\Lb^\top)^{-1}
 (\zb-\mb_i).
\end{align}
$\bmu_i$ is a trainable parameter that is supposed to simulate the class center $\bmu_i$. $\Lb$ is a real lower triangular matrix with positive diagonal entries, whose elements are also trainable. Based on the Cholesky decomposition, $\Lb\Lb^\top$ is guaranteed to be a symmetric positive-definite real matrix and is supposed to learn the covariance matrix $\bSigma$. Previous works in different manners have investigated the philosophy of designing Mahalanobis-distance-based scores. \citet{mahala} statistically computed the mean and variance matrix from features generated by a trained network. Such a design cannot leverage the benefit brought by training on auxiliary outliers. \citet{mmc} also proposed Max-Mahalanobis Center loss for better performance on adversarial robustness, although it adopted pre-defined constant values of $\bmu$ and $\bSigma$, which cannot be trained like our method. 

For in-distribution classification, the softmax cross-entropy loss will be computed upon our newly proposed distance based score $h_i(\zb)$ instead of $f_i(\zb)$ in Eqn.~\eqref{eq:ce}:
\begin{align}
\mathcal{L}_\text{sce}(\xb,\bm{1}_y)
=
-
\bm{1}_y^\top
\log
\big[\text{softmax}[h(\zb)]\big]
.
\end{align}
Compared with the simplified estimation in Eqn.~\eqref{eq:logistic}, our design provides a complete version of posterior probability estimation and prevents false likelihood. Our loss functions discussed below will utilize the estimated likelihood.


\subsection{A Probabilistic Loss Function}

In the last part, we modify the top layer of deep networks to estimate the likelihood under the assumption that the features of the penultimate layer are approximately class-conditional Gaussian distribution. The maximum likelihood value across all classes, \ie $\exp(\max\limits_{i} h_i)$,
can then be a good indicator to separate in-distribution and out-of-distribution features~\cite{maxlogit}. During training, we draw lessons from the classic Maximum Likelihood Estimation (MLE) of binary classification as the training loss:
\begin{align}
\label{eq:ice-ood}
&\cL_\text{ice-ood}(\tilde{\zb})
=
-
\log
\big[ 1-
\exp[\max\limits_{i} h_i(\tilde{\zb})]
\big],
\\
&\cL_\text{ice-id}(\zb, y)
=
-
\log
\big[
\exp[h_y(\zb)]
\big]
=
-h_y(\zb).
\label{eq:ice-id}
\end{align}
Eqn.~\eqref{eq:ice-id} drags in-distribution feature $\zb$ to class center and Eqn.~\eqref{eq:ice-ood} pushes outliers away. Their joint effect avoids distribution shift and results in the changing trend of features in Fig.~2. Note that, in Eqn.~\eqref{eq:ice-id}, instead of using the maximum likelihood $\exp(\max\limits_{i} h_i)$, we opt for the likelihood for the ground-truth $\exp[h_y(\zb)]$ to avoid potential contradictory gradient.
Finally, a weight parameter $\lambda$ is utilized to balance in-distribution learning and out-of-distribution learning:
\begin{align}
\label{eq:ice}
\textbf{ICE:} \ \
&\mathbb{E}_{(\xb,y)\sim\mathcal{D}_{in}}
\mathcal{L}_\text{ce}(\xb,\bm{1}_y) \
+
\\
&
\lambda \cdot 
\big[
\mathbb{E}_{(\xb,y)\sim\mathcal{D}_{in}}
\cL_{\text{ice-id}}(\xb, y)
+
\mathbb{E}_{\tilde{\xb}\sim\mathcal{D}_{out}^{oe}}
\cL_{\text{ice-ood}}(\tilde{\xb})
\big]
.
\notag
\end{align}

\subsection{A Versatile Solution in Practice}

Besides solving the three in-distribution incompatibility issues, ICE also possesses other benefit: 1) The ICE detector  $\exp[\max\limits_{i} h_i(\tilde{\zb})] \in (0,1]$ provides a direct estimation of detection confidence. Unlike previous methods, the learned maximum likelihood is explicitly trained by Eqn.~\eqref{eq:ice-id} and Eqn.~\eqref{eq:ice-ood}. In empirical evaluations, we will show that this learned confidence is well-calibrated. 2) The parameters of ICE are mostly learned through training, allowing us to easily add ICE into existing frameworks. Methods such as Mahalanobis~\cite{mahala} and ODIN~\cite{odin} need a second round of learning procedure after standard training. 3) Methods like Energy use multiple hyper-parameters to threshold the learned energy score. These hyper-parameters need to be exhaustively cross-validated and cannot be shared across different sets. In contrast, the only hyper-parameter of ICE is the $\lambda$ in Eqn.~\eqref{eq:ice}. 



\begin{table*}[t!]
	\caption{OOD detection Performance on various benchmarks, including different choices of models, training schedules, in-distribution sets ($\cD_{in}$), and out-of-distribution sets for testing ($\cD_{out}^{test}$). We compare our ICE with OE and Energy. Three detection metrics  (AUPR/AUROC/FPR95) are evaluated, where ICE consistently performs better. $\downarrow$ indicates that, as the metric decreases, the performance improves. $\uparrow$ indicates that, as the metric increases, the performance becomes better.}
	\label{table:vertical}
	\centering
	\resizebox{\textwidth}{!}{
		\begin{tabular}{ccccccc|ccc|ccc}
			\toprule
			& 
			& 
			& 
			& \multicolumn{3}{c}{AUPR $\uparrow$ }
			& \multicolumn{3}{c}{AUROC  $\uparrow$}
			& \multicolumn{3}{c}{FPR95  $\downarrow$}\\
			\cmidrule(r){5-7}
			\cmidrule(r){8-10}
			\cmidrule(r){11-13}
			$\cD_{in}$
			& Model
			& Training
			& $\cD_{out}^{test}$
			& OE     & Energy     & ICE(ours)
			& OE     & Energy     & ICE(ours)
			& OE     & Energy     & ICE(ours) \\
			\midrule
			\multirow{18}{*}{CIFAR10} 
			& \multirow{12}{*}{WideResNet}
			& \multirow{6}{*}{Fine-tune}
			& iSUN    
			& 98.79  & 98.92  & \textbf{99.31}
			& 99.15  & 99.29  & \textbf{99.45} 
			& 3.85  & 2.65  & \textbf{2.27}  \\
			& 
			& 
			& Places365    
			& 99.98  & 99.99  & \textbf{99.99}
			& 96.57  & \textbf{97.44}  & 97.16
			& 15.96  & \textbf{10.06}  & 11.73   \\
			& 
			& 
			& Textures    
			& 96.42  & 97.47  & \textbf{98.53}
			& 97.96  & 98.58  & \textbf{99.15} 
			& 10.27  & 5.31  & \textbf{4.37}  \\
			& 
			& 
			& LSUN    
			& 99.59  & 98.96  & \textbf{99.64}
			& 99.61  & 99.29  & \textbf{99.67} 
			& 1.81  & 2.53  & \textbf{1.46}  \\
			& 
			& 
			& SVHN    
			& 99.73  & 99.85  & \textbf{99.89}
			& 98.75  & 99.30  & \textbf{99.40} 
			& 3.84  & 2.19  & \textbf{1.94}  \\
			& 
			& 
			& CIFAR100    
			& 92.71  & 93.86  & \textbf{95.08}
			& 93.04  & 93.81  & \textbf{94.90} 
			& 30.42  & 28.24  & \textbf{23.23}  \\
			\cmidrule(r){3-13}
			&
			& \multirow{6}{*}{From-scratch}
			& iSUN    
			& 99.03  & 99.11  & \textbf{99.32}
			& 99.36  & 99.40  & \textbf{99.45} 
			& 2.29  & 2.31  & \textbf{2.30}   \\
			& 
			& 
			& Places365    
			& 99.99  & 99.99  & \textbf{99.99}
			& 97.48  & 97.58  & \textbf{97.68} 
			& 10.06  & \textbf{8.22}  & 9.97  \\
			& 
			& 
			& Textures    
			& 98.39  & 98.44  & \textbf{99.07}
			& 99.08  & 99.25  & \textbf{99.51} 
			& 3.55  & 2.49  & \textbf{1.96}  \\
			& 
			& 
			& LSUN    
			& 99.32  & 99.15  & \textbf{99.47}
			& 99.45  & 99.50  & \textbf{99.62} 
			& 2.11  & 1.25  & \textbf{1.20}  \\
			& 
			& 
			& SVHN    
			& 99.91  & \textbf{99.95}  & 99.91 
			& 99.48  & \textbf{99.76}  & 99.58 
			& 2.15  & \textbf{0.79}  & 1.26   \\
			& 
			& 
			& CIFAR100    
			& 94.95  & 94.65  & \textbf{95.38}
			& 94.79  & 94.44  & \textbf{95.13} 
			& 27.10  & 26.15  & \textbf{22.36}  \\
			\cmidrule(r){2-13}
			& \multirow{6}{*}{DenseNet}
			& \multirow{6}{*}{From-scratch}
			& iSUN    
			& 97.05  & 98.61  & \textbf{98.93}
			& 98.14 & 99.14  & \textbf{99.15} 
			& 7.09  & 3.54  & \textbf{3.52}  \\
			& 
			& 
			& Places365    
			& 99.98  & 99.98  & \textbf{99.98}
			& 96.19  & 96.14  & \textbf{96.21} 
			& \textbf{15.24}  & 18.15  & 20.25   \\
			& 
			& 
			& Textures    
			& 96.40  & \textbf{97.89} & 97.62 
			& 98.14  & \textbf{98.91}  & 98.46  
			& 8.36   & \textbf{4.02}  & 9.05   \\
			& 
			& 
			& LSUN    
			& 98.32  & 98.10  & \textbf{98.97}
			& 98.70  & 98.88  & \textbf{99.04} 
			& 5.11  &  3.59 & \textbf{3.79}  \\
			& 
			& 
			& SVHN    
			& 99.44  & 99.83  & \textbf{99.89}
			& 97.66  & 99.24  & \textbf{99.28} 
			& 7.10  & 3.01  & \textbf{2.98}   \\
			& 
			& 
			& CIFAR100    
			& 91.18  & 90.91  & \textbf{91.90}
			& 91.40 & 90.37  & \textbf{91.51} 
			&  41.13  & 44.80  & \textbf{40.44}  \\
			\midrule
			\multirow{5}{*}{CIFAR100} 
			& \multirow{5}{*}{WideResNet}
			& \multirow{5}{*}{From-scratch}
			& iSUN    
			& 83.61  & \textbf{89.39}  & 85.34 
			& 88.40  & \textbf{91.70}  & 88.45 
			& 36.55  & \textbf{29.88}  & 39.05  \\
			& 
			& 
			& Places365    
			& 99.91  & 99.92  & \textbf{99.92}
			& 87.25  & 87.62  & \textbf{87.55} 
			& \textbf{44.89}  & 45.42  & 49.00  \\
			& 
			& 
			& Textures    
			& 78.32  & 79.92  & \textbf{81.15}
			& 88.23  & 88.41  & \textbf{89.68} 
			& 39.84  & 43.57  & \textbf{38.95}  \\
			& 
			& 
			& LSUN    
			& \textbf{94.11}  & 89.51  & 92.87 
			& 94.02  & 92.53  & \textbf{94.15} 
			& \textbf{15.35}  & 26.84  & 25.37  \\
			& 
			& 
			& SVHN    
			& 98.92  & 98.71  & \textbf{98.97}
			& 94.49  & 93.41  & \textbf{94.71} 
			& 22.57  & 23.65  & \textbf{22.41}  \\
			\bottomrule
		\end{tabular}
	}
\end{table*}

\section{Experiment}

In this section, we verify the effectiveness of ICE on a wide range of OOD evaluation benchmarks.

\subsection{Experimental Setup}

\textbf{Dataset.} For the in-distribution set, we choose the standard CIFAR10 and CIFAR100~\cite{cifar} as our major verification target. For out-of-distribution set, we adopt several commonly-used benchmarks, including Textures~\cite{texture}, SVHN~\cite{svhn}, Places365~\cite{places365}, LSUN~\cite{lsun}, and iSUN~\cite{isun}. We also use CIFAR100 as an OOD source to evaluate models learned on CIFAR10 and vice versa. We choose the 80 Million Tiny Images~\cite{80mn} as the $\cD_{out}^{oe}$.

\noindent\textbf{Training.} Following \citet{oe}, we test two training protocols. Fine-tune: We initialize the model with a pre-trained checkpoint on the in-distribution set and then fine-tune the model with ten epochs. We adopt a cosine decay learning rate schedule with the initial value of $0.01$. From-scratch: We train deep networks with both $\cD_{in}^{train}$ and $\cD_{out}^{oe}$ for 100 epochs. The initial learning rate is set to $0.1$ with the commonly-used stair-wise decay learning rate schedule. For both protocols, the batch size for $\cD_{in}^{train}$ is set to $128$, and $\cD_{out}^{oe}$ to 256.

\noindent\textbf{Evaluation Metric.} We adopt three mostly commonly-used OOD detection metrics for evaluation: the area under the precision-recall curve (AUPR), the area under the receiver operating characteristic curve (AUROC), and the false positive rate at $95\%$ true positive rate (FPR95).

\subsection{Outlier Detection Performance}

We present our main results in Table.~\ref{table:vertical}. For each experimental setup, we compare our ICE with OE~\cite{oe} and Energy~\cite{energy}. Our ICE outperforms the other two counterparts on most benchmarks. For CIFAR10 as $\cD_{in}$, we first present results on the model of WideResNet-34-10. Both fine-tune and from-scratch schedules achieve promising results for the three algorithms, while the from-scratch schedule slightly but consistently outperforms fine-tune schedule. This aligns with the observations by \citet{oe}. To verify the extensibility of our method, we also present results on the DenseNet-40-12 model with the from-scratch schedule. Again, ICE exhibits generally better detection capability. Note that, for most choices of $\cD_{out}^{test}$, all the three methods have achieved very high scores, leaving only very little room for improvement. In contrast, for the hardest case where CIFAR100 serves as  $\cD_{out}^{test}$, the improvement brought by ICE becomes more significant. For the WideResNet model learned by fine-tune schedule, ICE achieves $95.08\%$ AUPR, $94.90\%$ AUROC, and $23.23\%$ FPR95, with the improvement of $2.37\%/1.86\%/7.19\%$ over OE. For CIFAR100 as $\cD_{in}$, we test WideResNet with the from-scratch schedule. Again, ICE achieves state-of-the-art results. In the case of 
SVHN being $\cD_{out}^{test}$, ICE achieves $98.97\%$ AUPR, $94.71\%$ AUROC, and $22.41\%$ FPR95.

\subsection{In-Distribution Accuracy}

ICE is designed for better in-distribution compatibility. Therefore, in this part, we compare its in-distribution accuracy with other baselines, including those without training on outliers. In Table.~\ref{table:horizontal}, we show the average detection scores across iSUN, Places365, Textures, LSUN, and SVHN. OE and Energy with the from-scratch schedule achieve superior detection performance and greatly improve detection performance over non-training detection algorithms such as ODIN and Mahalanobis but degrade the in-distribution accuracy. OE and Energy with fine-tune schedules achieve relatively better in-distribution accuracy than non-training detection algorithms. The small learning rate of fine-tune schedule prevents the degradation of in-distribution accuracy. However, the fine-tune schedule cannot reach the same detection performance level as the from-scratch schedule.

In contrast, ICE accomplishes evident higher accuracy than other baselines, either under fine-tune or from-scratch schedules. The objective of ICE effectively utilizes the auxiliary examples from $\cD_{out}^{oe}$ to provide neural network with new information and thus help its generalization. The refined in-distribution compatibility concentrates in-distribution features around the class center, endows in-distribution classification with high confidence, and thus further helps detect outliers with low confidence. Unlike OE and Energy, ICE achieves better in-distribution accuracy with a from-scratch schedule than fine-tune schedule, indicating its intrinsic built-in in-distribution compatibility.

\begin{table}[t!]
	\caption{The OOD detection performance and in-distribution accuracy for different methods. $\dagger$ represents models trained with fine-tune schedule. $\ddagger$ represents models trained with the from-scratch schedule.}
	\label{table:horizontal}
	\centering
	\resizebox{\linewidth}{!}{
		\begin{tabular}{cccccc}
			\toprule
			  $\cD_{in}$
			& Method
			& AUPR $\uparrow$ 
			& AUROC  $\uparrow$
			& FPR95  $\downarrow$ 
			& \tabincell{c}{In-dist\\Accuracy} $\uparrow$ \\
			\midrule
			\multirow{9}{*}{\tabincell{c}{CIFAR\\-10}} 
			& MSP
			& 97.88  & 90.82  & 56.03 & 95.11 \\
			& ODIN
			& 97.39  & 90.39  & 37.53 & 95.11 \\
			& Mahalanobis
			& 98.47  & 93.27  & 35.97 & 95.11 \\
			& OE$\dagger$
			& 98.90  & 98.40 & 7.14 & 95.23 \\
			& Energy$\dagger$
			& 99.03  & 98.78  & 4.54 & 95.88 \\
			& ICE(ours)$\dagger$
			& 99.47  & 98.96  & 4.35 & 96.02 \\
			& OE$\ddagger$
			& 99.32  & 98.97  & 4.03 & 94.88 \\
			& Energy$\ddagger$
			& 99.32  & 99.10  & 4.12 & 94.13 \\
			& ICE(ours)$\ddagger$
			& \textbf{99.55}  & \textbf{99.16}  & \textbf{3.33} & \textbf{96.38} \\
            \midrule
			\multirow{6}{*}{\tabincell{c}{CIFAR\\-100}} 
			& MSP
			& 93.90  & 75.56  & 80.01 & 76.01 \\
			& ODIN
			& 93.94  & 76.55  & 75.17 & 76.01 \\
			& Mahalanobis
			& 95.22  & 81.74  & 60.60 & 76.01 \\
			& OE$\ddagger$
			& 90.97  & 90.47  & \textbf{31.84} & 75.82 \\
			& Energy$\ddagger$
			& 91.49  & 90.73  & 33.87 & 75.75 \\
			& ICE(ours)$\ddagger$
			& \textbf{91.65}  & \textbf{90.91}  & 34.95 & \textbf{77.74} \\
			\bottomrule
		\end{tabular}
	}
\end{table}

\begin{figure*}[t!]
\centering
\resizebox{0.95\linewidth}{!}{
		\subfigure[Plain Training.]{
			\includegraphics[width=0.20\linewidth]{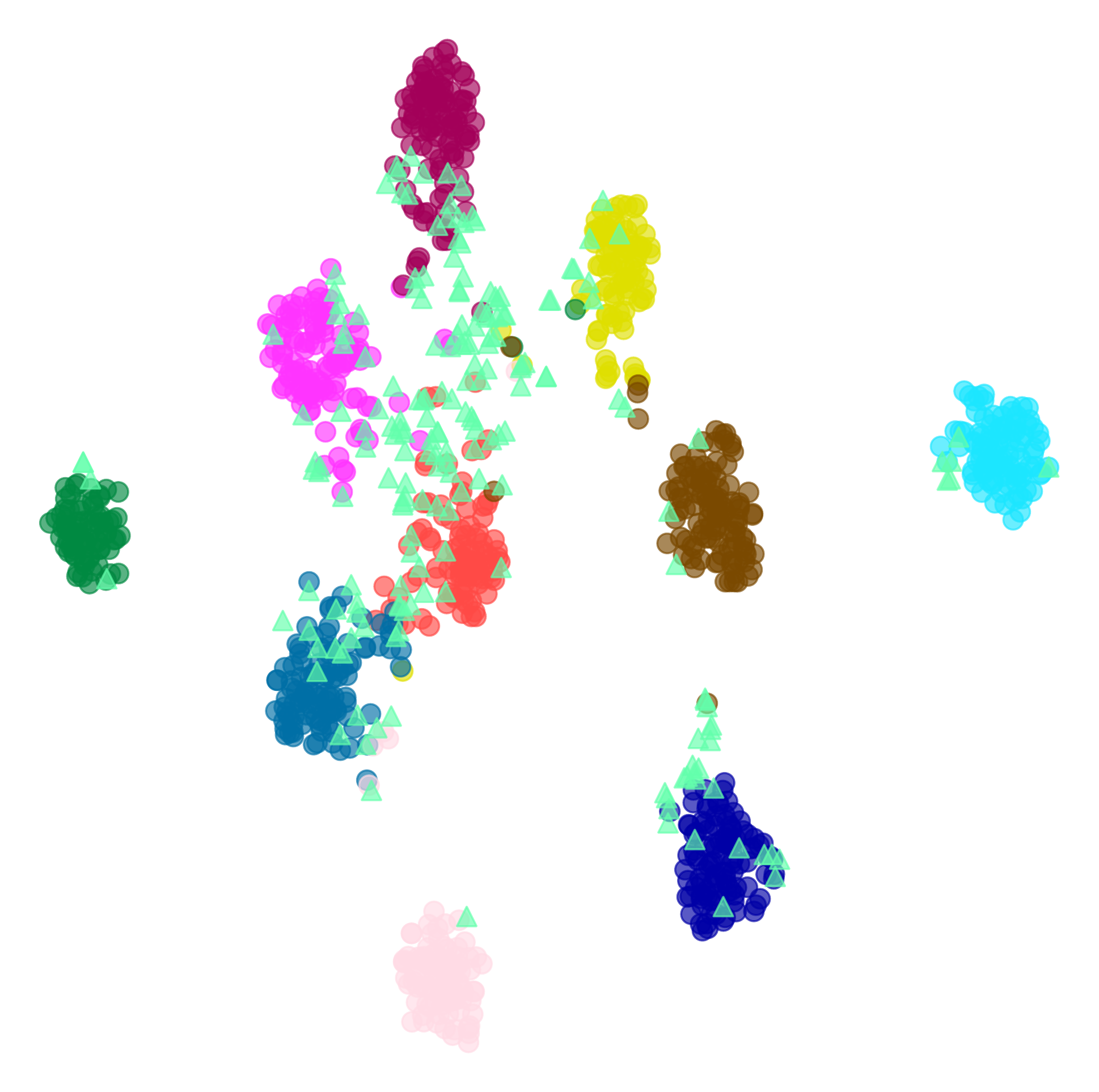}}
        \quad
		\subfigure[Outlier Exposure.]{
			\includegraphics[width=0.20\linewidth]{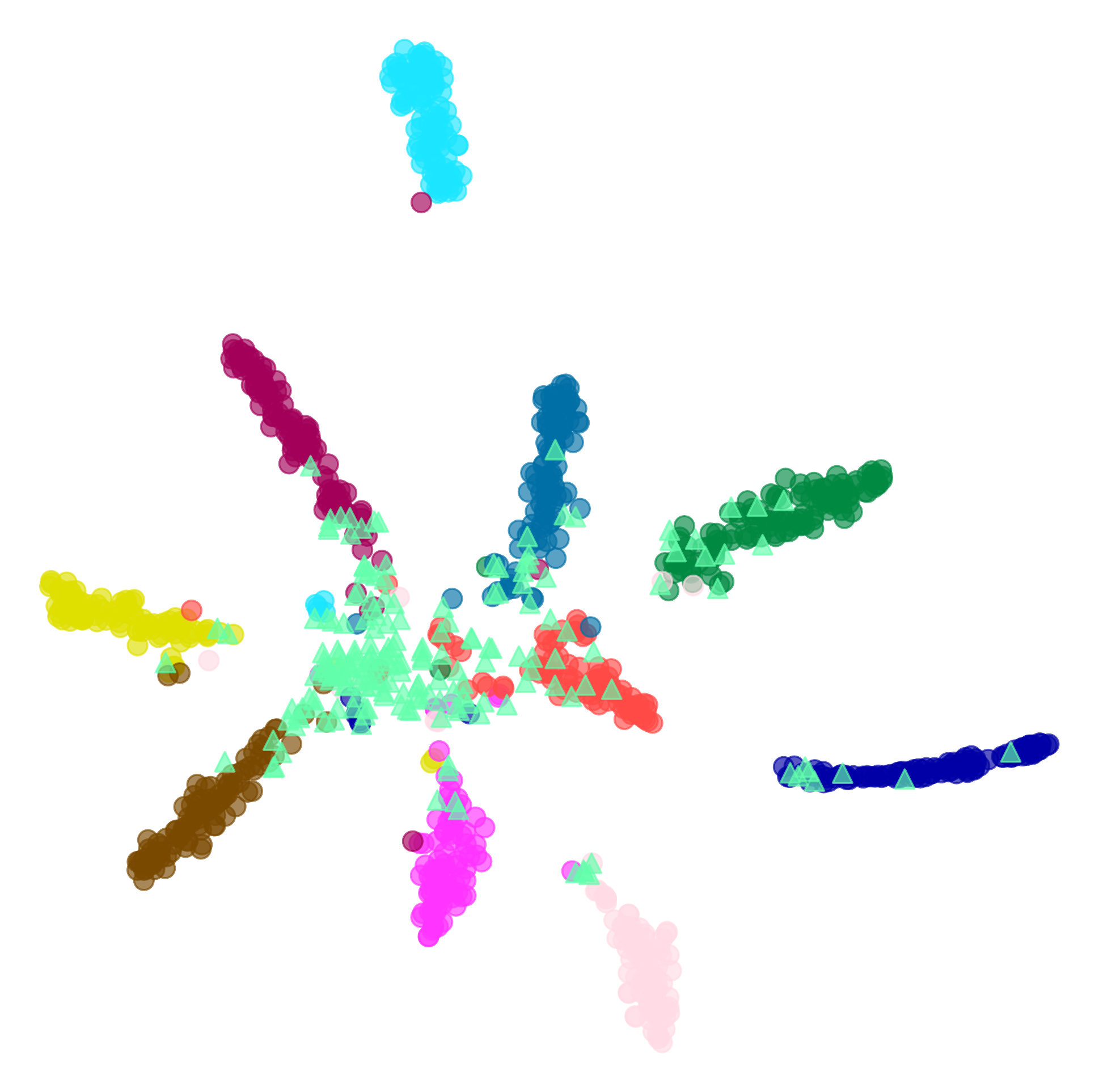}}
		\quad
		\subfigure[Energy Score.]{
			\includegraphics[width=0.20\linewidth]{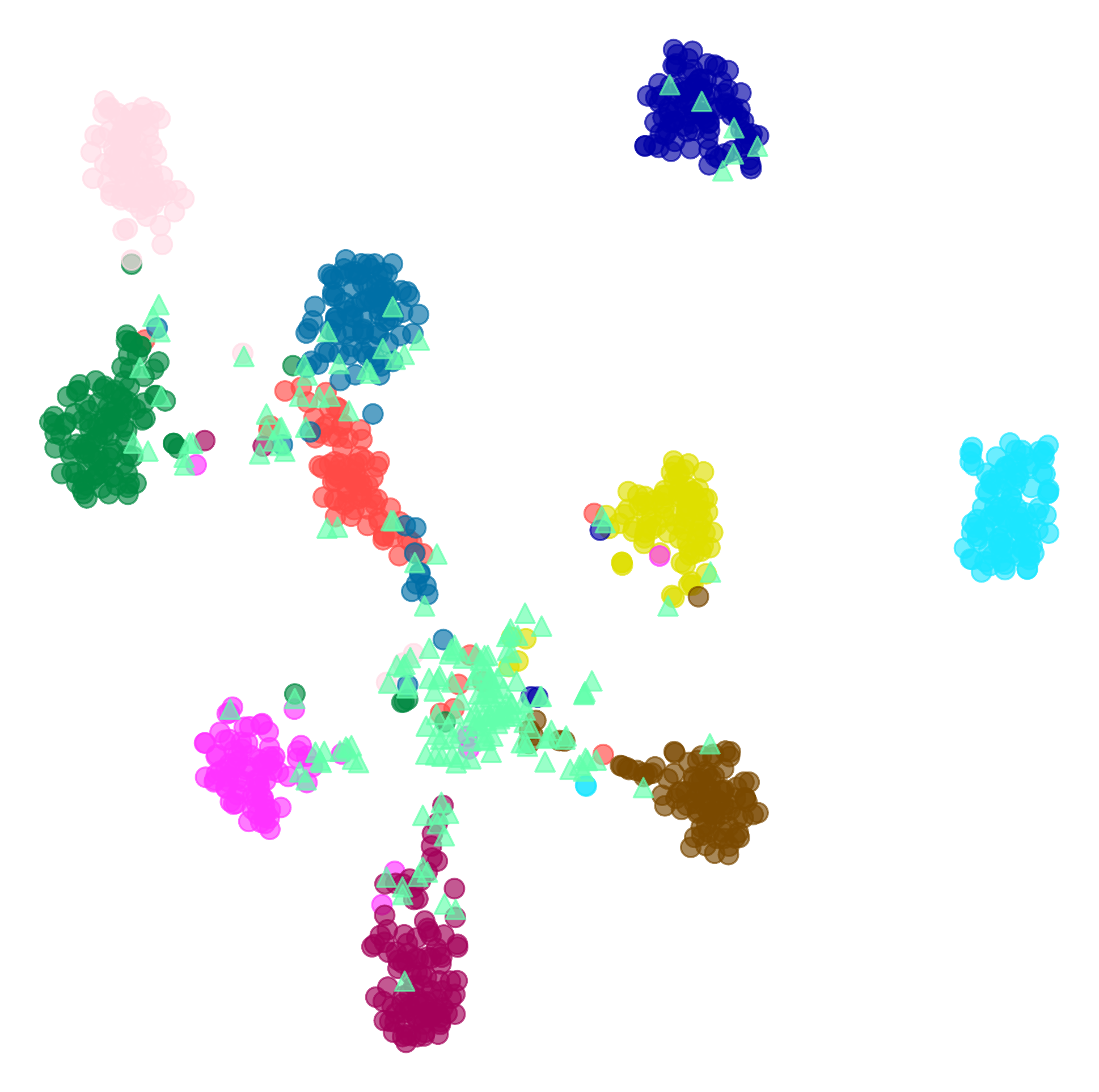}}
        \quad
		\subfigure[ICE (ours).]{
			\includegraphics[width=0.20\linewidth]{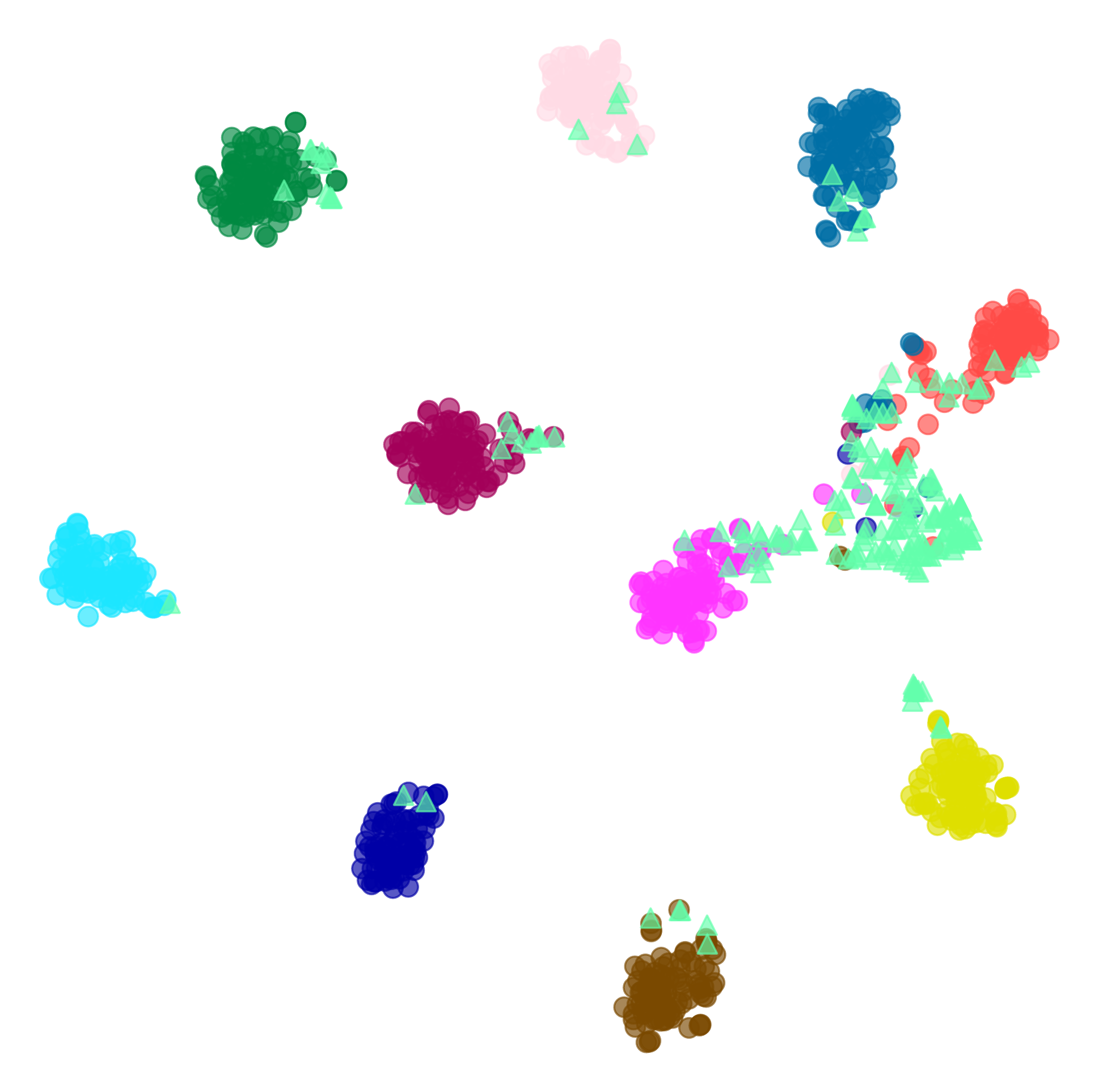}}
}
		\caption{Feature visualization with t-SNE algorithm. We present three OE methods and Plain Training. ICE learns compact and discriminating features. {\color{teal}{$\blacktriangle$}} represents OOD examples (CIFAR100), and the rest examples are in-distribution (CIFAR10).}
		\label{fig:tsne}
\end{figure*}

\subsection{Feature Distribution}

To get a straightforward insight into how different outlier training methods influence the distribution, we visualize the features generated by the penultimate layer of deep networks, \ie $\zb$ and $\tilde{\zb}$, with the t-SNE algorithm. In our visualization, the t-SNE mapping is first learned on all in-distribution samples. Then, we use the learned mapping to transform outliers. As shown in Fig.~\ref{fig:tsne}, the Plain Training method without any learning on outliers exhibit poor OOD feature distribution. The OOD features are heavily mixed with in-distribution ones. Standard Outlier Exposure squeezes OOD features in the space center and pushes in-distribution ones into diverse directions. This aligns our visualization on a 2-dimension case in Fig.~\ref{fig:ds}. Due to the effect of OE, the original clustered in-distribution features are transformed into slender ones. Energy maintains the cluster shape of in-distribution features. However, like the OE algorithm, the outlier features are close to in-distribution ones, which increases the difficulty of discriminating them. ICE, on the other hand, resumes the most characteristics of in-distribution features in Plain Training. The in-distribution clusters are tightly bounded and scatter from each other. Meanwhile, outliers are mostly constrained in a local space.

\subsection{Ablation Study}

In this part, we conduct ablation studies on each component of ICE. First of all, ICE modifies the top design of WideResNet. We dub this modified model as WideResNet(G). To evaluate the effect of this design, we test Energy algorithm on  WideResNet(G), which addresses the false likelihood problem but still has the issue of contradictory gradient. As shown in Table~\ref{table:ablation}, Energy method with WideResNet(G) perform slightly better than the baseline with WideResNet. Then, the second counterpart we want to compare is applying the Sigmoid Binary Cross-Entropy (BCE) loss onto WideResNet. Such design avoids the contradictory gradient issue yet still has false likelihood and distribution shift problems. The improvement is still observable but limited. We also test eliminating the $\cL_{\text{energy-id}}$ in Eqn.~\eqref{eq:ice} (ICE$^-$), which prevents contradictory gradient and false likelihood but not distribution shift. The detection performance is heavily degraded, indicating the importance of balancing $\cL_{\text{energy-ood}}$ with $\cL_{\text{energy-ood}}$ in Eqn.~\eqref{eq:ice}. The above ablation studies show that the individual improvement of each component is not as significant as the total improvement. The superior performance of ICE is achieved by the collective effect of all components.

\begin{table}[h!]
	\caption{Ablation studies on various components of ICE. }
	\label{table:ablation}
	\centering
	\resizebox{\linewidth}{!}{
		\begin{tabular}{cccccc}
			\toprule
			  Loss Function
			& Model
			& AUPR $\uparrow$ 
			& AUROC  $\uparrow$
			& FPR95  $\downarrow$ 
			& \tabincell{c}{In-dist\\Accuracy} $\uparrow$ \\
			\midrule
			Energy & WideResNet
			& 99.32  & 99.10  & 4.12 & 94.13 \\
			Energy & WideResNet(G)
			& 99.37  & 99.14  & 4.01 & 93.95 \\
			BCE & WideResNet
			& 99.40  & 99.02  & 4.48 & 95.81 \\
			ICE$^-$ & WideResNet(G)
			& 83.62  & 74.03  & 35.58 & 90.44 \\
			ICE & WideResNet(G)
			& \textbf{99.55}  & \textbf{99.16}  & \textbf{3.33} & \textbf{96.38} \\
			\bottomrule
		\end{tabular}
	}
\end{table}

\begin{figure}[h!]
\centering
\includegraphics[width=0.95\linewidth]{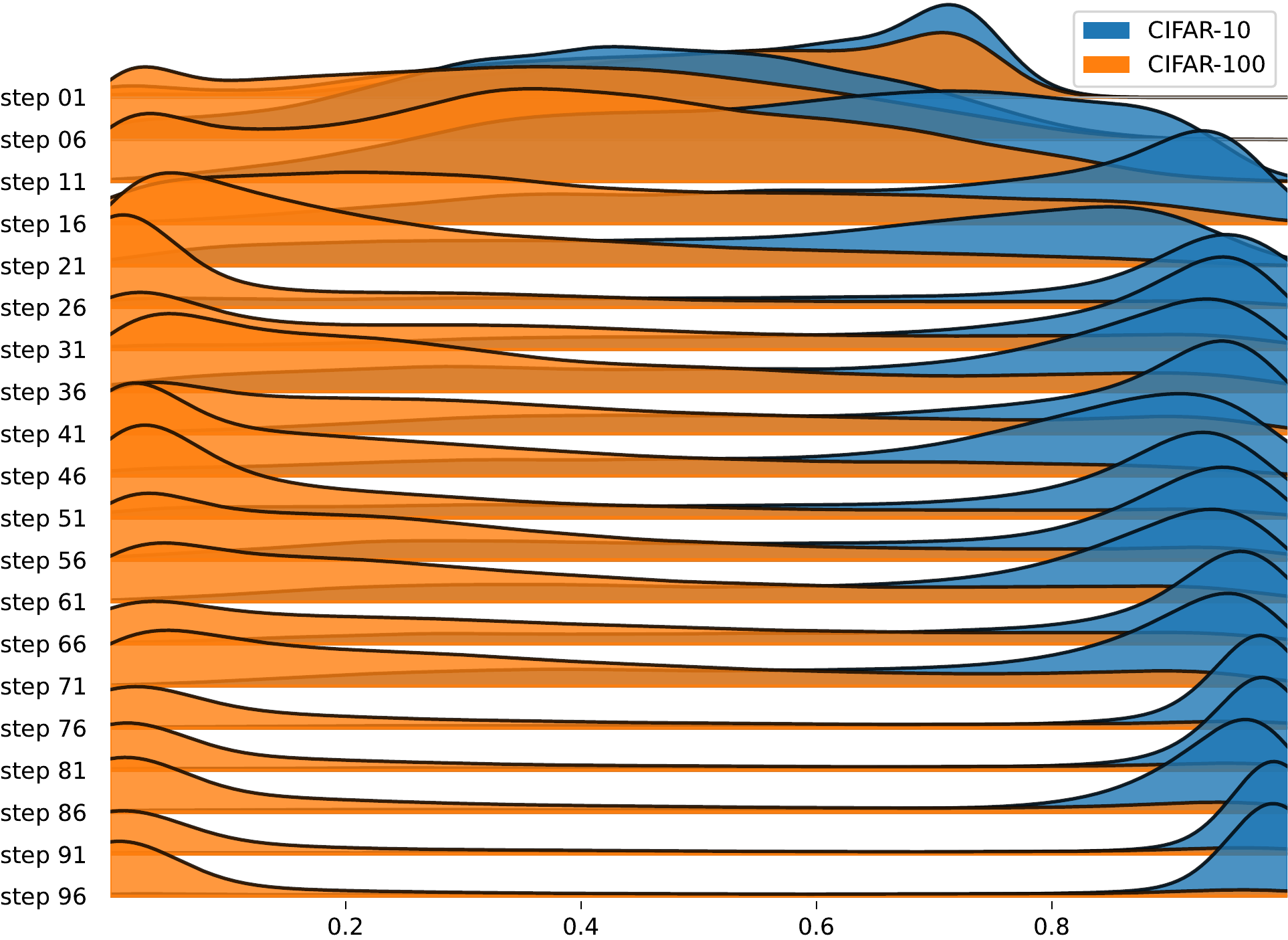}
\caption{On each training epoch, we plot the distribution of detection confidence for in-distribution set (CIFAR10) and out-of-distribution set (CIFAR100).}
\label{fig:joy}
\end{figure}

\subsection{Confidence Estimation}
As discussed before, unlike OE and Energy, our method provides the capability of estimating the $(0,1]$ probability of whether the upcoming input is in-distribution or not. Here we present a direct visualization of our in-distribution confidence indicator of $\exp[\max\limits_{i} h_i(\tilde{\zb})]$. In Fig.~\ref{fig:joy}, we plot the distribution of $\exp[\max\limits_{i} h_i(\zb)]$ (CIFAR10) and $\exp[\max\limits_{i} h_i(\tilde{\zb})]$ (CIFAR100) against the training epochs. As the training progresses, the two distributions gradually separate from each other and eventually concentrate on their ground-truth value. Moreover, the two distributions both formulate a typical pattern of long-tail distribution, indicating that the majority of the samples are trained to their desirable category.

\section{Conclusion}

In this paper, we analyzed existing Outlier Exposure methods and demonstrated that those methods are detrimental to in-distribution accuracy due to three factors: \textit{contradictory gradient}, \textit{false likelihood}, and \textit{distribution shift}. We thereafter proposed a novel Outlier Exposure method, namely ICE, to address the three defects. ICE replaces a conventional linear discriminator with a Gaussian-like discriminator to prevent false likelihood. Then, the likelihood score generated by the Gaussian-like discriminator is trained via a loss function, enlightened by the classic MLE solution for binary classification. The loss function can prevent distribution shift and does not yield contradictory gradients. ICE achieves \textit{sota} results on multiple OOD detection benchmarks through the collective effects of the above-designed components. Meanwhile, ICE improves in-distribution accuracy by learning from additional OOD features.


\bibliography{a-reference}
\clearpage

\section{Appendix}

\subsection{Example Details for Fig.~2\&3}

We describe the detail of generating samples for the 2-dimensional binary classification task in Fig.~2\&3. We randomly draw samples from $p(\xb|c_1) \sim \cN ((\mu,0),\bI)$ and $p(\xb|c_2) \sim\cN ((-\mu,0),\bI)$ separately. $\bI$ is the identity matrix. A drawn sample is considered as an in-distribution one when $\max( (p(\xb|c_1), p(\xb|c_2)) > \zeta$, otherwise out-of-distribution. $\zeta$ is an arbitrary thresh-hold

\subsection{Balance Weight $\lambda$}

\begin{align}
\textbf{OE:} \ \ 
\underbrace{
\EE_{(\xb,y)\sim\cD_{in}}
\cL_\text{sce}(\xb,\bm{1}_y)
}_{\text{in-distribution risk}}
+
\lambda \cdot
\underbrace{
\mathbb{E}_{\tilde{\xb}\sim\mathcal{D}_{out}^{oe}}
\mathcal{L}_\text{sce}(\tilde{\xb}, \ub)
}_{\text{outlier exposure}}
.
\end{align}

\begin{align}
\textbf{Energy:}  \ \ 
&\mathbb{E}_{(\xb,y)\sim\mathcal{D}_{in}}
\mathcal{L}_\text{sce}(\xb,\bm{1}_y) \
+
\\
&
\lambda \cdot 
\big[
\mathbb{E}_{(\xb)\sim\mathcal{D}_{in}}
-
\cL_{\text{energy}}(\xb) 
+
\mathbb{E}_{\tilde{\xb}\sim\mathcal{D}_{out}^{oe}}
\cL_{\text{energy}}(\tilde{\xb})
\big]
.
\notag
\end{align}

\begin{align}
\label{eq:ice}
\textbf{ICE:} \ \
&\mathbb{E}_{(\xb,y)\sim\mathcal{D}_{in}}
\mathcal{L}_\text{ce}(\xb,\bm{1}_y) \
+
\\
&
\lambda \cdot 
\big[
\mathbb{E}_{(\xb,y)\sim\mathcal{D}_{in}}
\cL_{\text{ice-id}}(\xb, y)
+
\mathbb{E}_{\tilde{\xb}\sim\mathcal{D}_{out}^{oe}}
\cL_{\text{ice-ood}}(\tilde{\xb})
\big]
.
\notag
\end{align}

For OE, Energy, and our ICE, there is a parameter $\lambda$ balancing the in-distribution learning and out-of-distribution learning. Here we cross-validated its value for better understanding. OE sets the default value of $\lambda$ to $0.5$, Energy sets the default value of $\lambda$ to $0.1$. ICE sets $1.0$. We test scaling the default $\lambda$ of each method with an extra parameter $\gamma$.

\begin{table}[h!]
	\caption{Ablation studies on $\lambda$. }
	\centering
	\resizebox{0.95\linewidth}{!}{
		\begin{tabular}{ccccccc}
			\toprule
			Metric
			& Method
			& $\gamma=1.0$
			& $\gamma=3.0$
			& $\gamma=5.0$
			& $\gamma=7.0$
			& $\gamma=9.0$ \\
			\midrule
			\multirow{3}{*}{AUPR} 
			& OE
			& 99.32 & 99.25 & 99.17 & 99.21 & 99.02  \\
			& Energy
			& 99.32 & NaN & NaN & NaN & NaN  \\
			& ICE
			& 99.55 & 99.46 & 99.21 & 99.37 & 99.14  \\
			\midrule
			\multirow{3}{*}{AUROC} 
			& OE
			& 98.97 & 98.81 & 98.65 & 98.69 & 97.84  \\
			& Energy
			& 99.10 & NaN & NaN & NaN & NaN    \\
			& ICE
			& 99.19 & 99.17 & 99.08 & 98.85 & 98.61  \\
			\midrule
			\multirow{3}{*}{FPR95} 
			& OE
			& 4.03 & 4.25 & 4.98 & 5.12 & 6.04  \\
			& Energy
			& 4.12 & NaN & NaN & NaN & NaN    \\
			& ICE
			& 3.33 & 3.59 & 3.82 & 3.90 & 4.07  \\
			\midrule
			\multirow{3}{*}{In-dist} 
			& OE
			& 94.88 & 94.92 & 94.51 & 94.33 & 94.10  \\
			& Energy
			& 94.13 & NaN & NaN & NaN & NaN     \\
			& ICE
			& 96.38 & 96.27 & 96.11 & 95.82 & 95.13  \\
			\bottomrule
		\end{tabular}
	}
\end{table}

The table above shows that our ICE can still achieve desirable results on larger weight parameters. OE also shows a stable performance for larger $\lambda$. However, Energy cannot be trained with large $\lambda$. Once the weight parameter is large than three times of the optimal value, the training would simply result in NaN loss value and cannot achieve a reasonable performance. 

\subsection{Logit Distribution}
In Fig.~3, we have visualized that OE methods may shift the distribution of in-distribution features for 2-dimension case. In Fig.~6, we plot the maximum logit value of neural network to see how OE methods influence the feature distribution of networks. The extra learning of OE and Energy shifts the maximum logit to larger values. This is a consequence of distribution shift, which verifies our analyses before.

\begin{figure}[h!]
\centering
\includegraphics[width=0.95\linewidth]{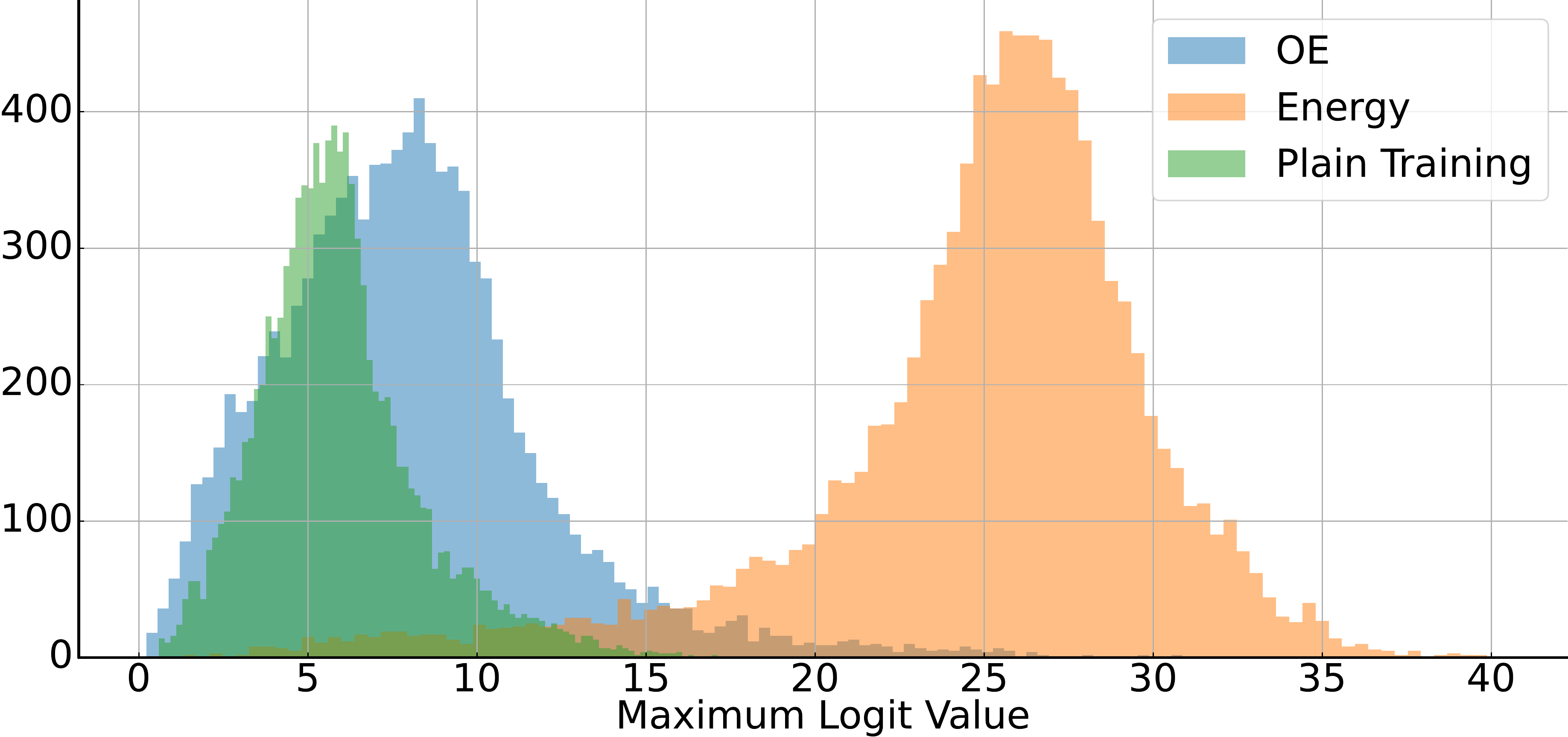}
\caption{Distribution of maximum logit value of neural network for different trainig objective.}
\end{figure}

\subsection{Extra Feature Visualization}
In Fig.~4, we visualize features with t-SNE algorithm, where OOD examples are from the CIFAR100 set. In Fig.~7, we present the case where SVHN serves as the OOD set.

\begin{figure*}[t!]
\centering
\resizebox{0.95\linewidth}{!}{
		\subfigure[Plain Training.]{
			\includegraphics[width=0.20\linewidth]{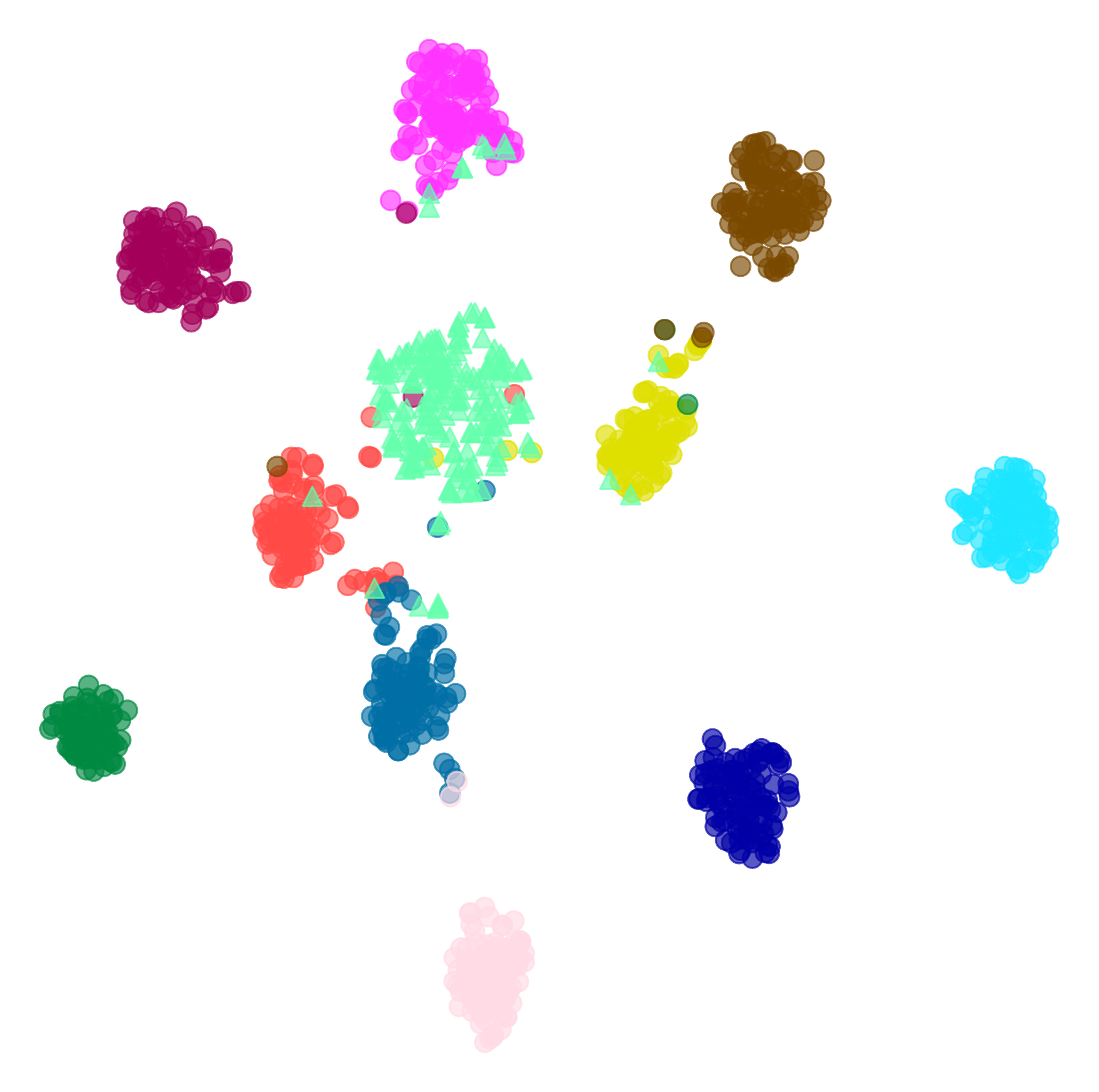}}
        \quad
		\subfigure[Outlier Exposure.]{
			\includegraphics[width=0.20\linewidth]{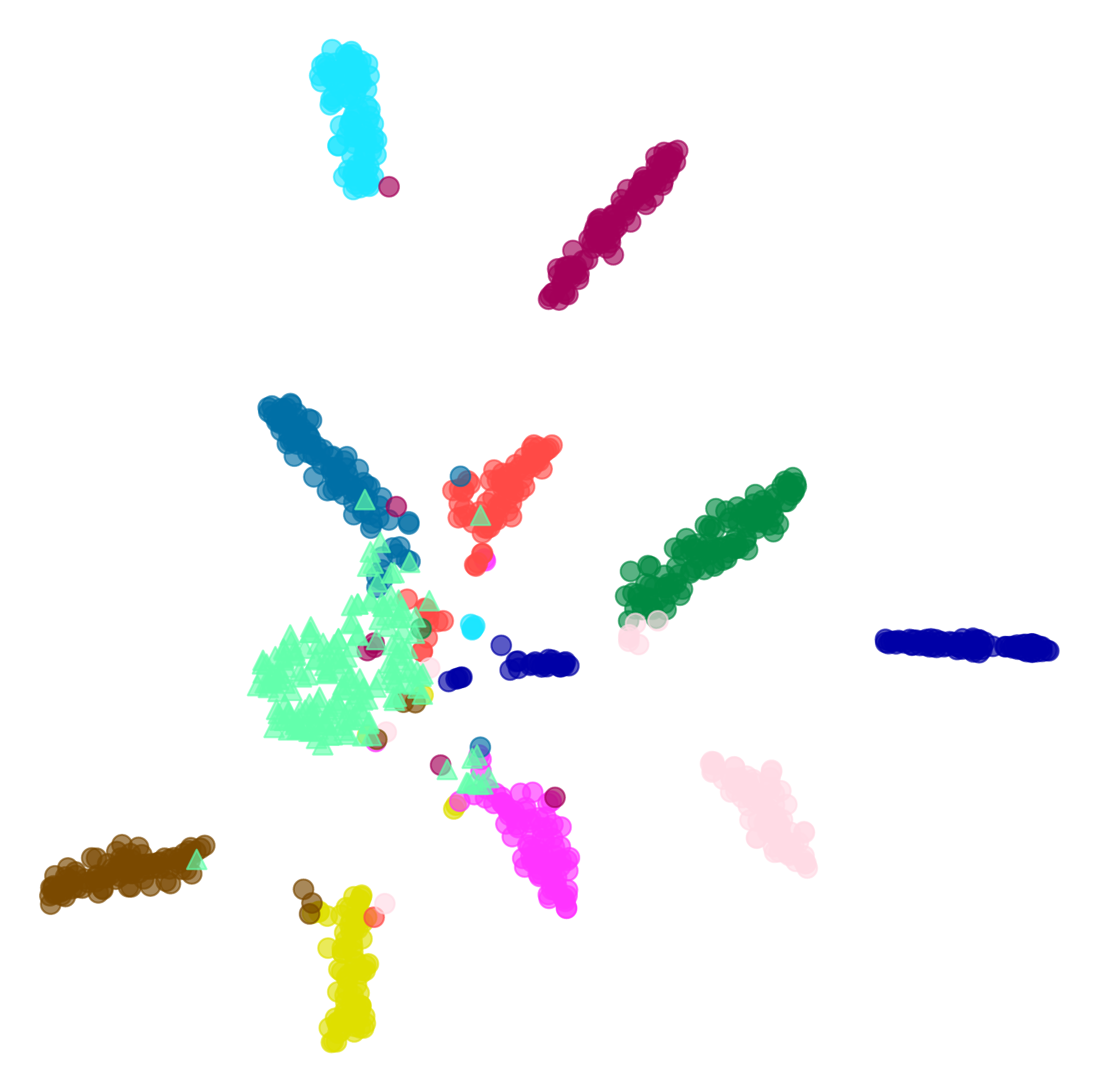}}
		\quad
		\subfigure[Energy Score.]{
			\includegraphics[width=0.20\linewidth]{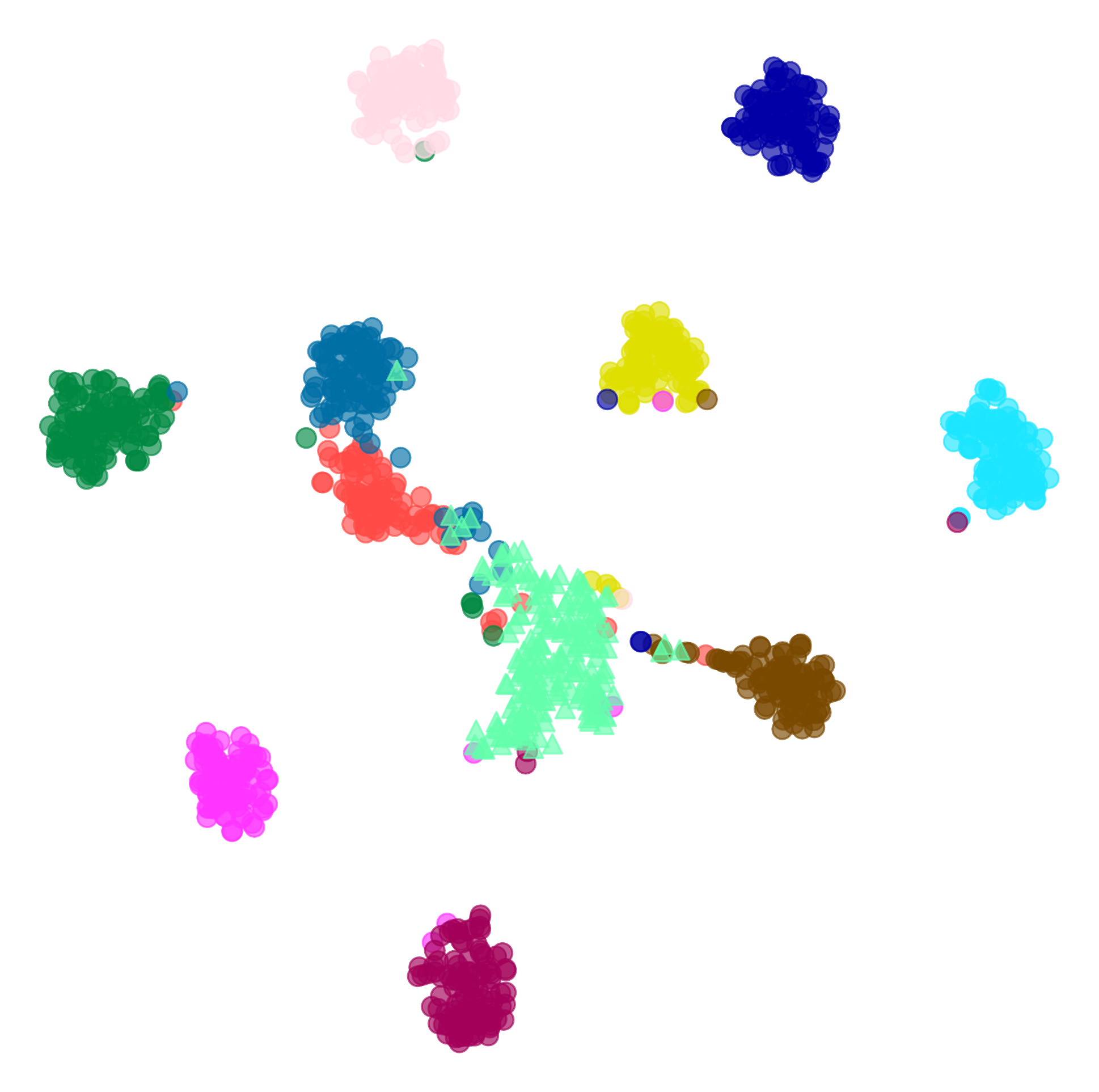}}
        \quad
		\subfigure[ICE (ours).]{
			\includegraphics[width=0.20\linewidth]{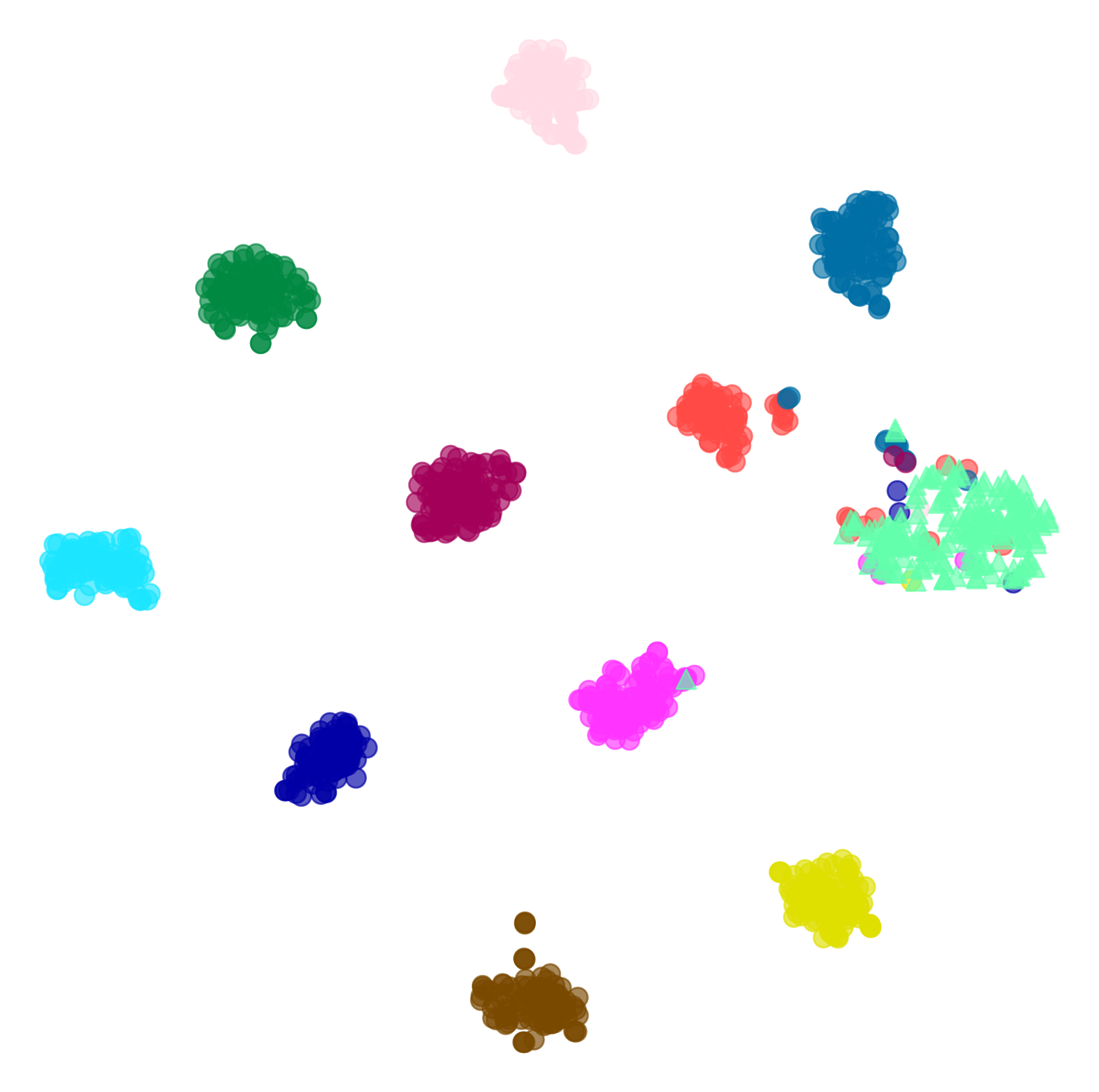}}
}
		\caption{Feature visualization with t-SNE algorithm. We present three OE methods and Plain Training. ICE learns compact and discriminating features. {\color{teal}{$\blacktriangle$}} represents OOD examples (SVHN), and the rest examples are in-distribution (CIFAR10).}
		\label{fig:tsne-svhn}
\end{figure*}

\end{document}